\documentclass[authoryear]{elsarticle}
\usepackage{rotating}
\usepackage{ifpdf}

\ifpdf
\usepackage[%
  pdftitle={Instructions for use of the document class
    elsart},%
  pdfauthor={Simon Pepping},%
  pdfsubject={The preprint document class elsart},%
  pdfkeywords={instructions for use, elsart, document class},%
  pdfstartview=FitH,%
  bookmarks=true,%
  bookmarksopen=true,%
  breaklinks=true,%
  colorlinks=true,%
  linkcolor=blue,anchorcolor=blue,%
  citecolor=blue,filecolor=blue,%
  menucolor=blue,pagecolor=blue,%
  urlcolor=blue]{hyperref}
\else
\usepackage[%
  breaklinks=true,%
  colorlinks=true,%
  linkcolor=blue,anchorcolor=blue,%
  citecolor=blue,filecolor=blue,%
  menucolor=blue,pagecolor=blue,%
  urlcolor=blue]{hyperref}
\fi

\makeatletter
\def\elsartstyle{%
    \def\normalsize{\@setfontsize\normalsize\@xiipt{14.5}}
    \def\small{\@setfontsize\small\@xipt{13.6}}
    \let\footnotesize=\small
    \def\large{\@setfontsize\large\@xivpt{18}}
    \def\Large{\@setfontsize\Large\@xviipt{22}}
    \skip\@mpfootins = 18\p@ \@plus 2\p@
    \normalsize
}
\@ifundefined{square}{}{}
\makeatother

\usepackage{tikz}
\usetikzlibrary{decorations.pathreplacing,calc}

\usepackage{algpseudocode}% http://ctan.org/pkg/algorithmicx
\usepackage{algorithm}% http://ctan.org/pkg/algorithm
\usepackage{subfig}
\usepackage{multirow}
\usepackage{amsfonts}
\usepackage{amsmath}
\usepackage{amsthm}
\usepackage{amssymb}
\usepackage{mathtools}
\usepackage{breqn}
\usepackage{rotating}
\usepackage{bruza}
\usepackage{listings}
\usepackage{csvsimple}
\usepackage{stmaryrd}
\usepackage{lineno}

\theoremstyle{definition}
\newtheorem{definition}{Definition}[section]

\begin{document}
%\linenumbers
\begin{frontmatter}
\title{Modelling contextuality by probabilistic programs with hypergraph semantics}
\author[qutis]{Peter D. Bruza}%\corref{cor1}} 
\ead{p.bruza@qut.edu.au}
%\cortext[cor1]{Corresponding author}
\address[qutis]{School of Information Systems\\
Queensland University of Technology \\
GPO Box 2434\\
Brisbane 4001\\
Australia}

\begin{abstract}
%what we did
%why we did it
Models of a phenomenon are often developed by examining it under different experimental conditions, or measurement contexts.
The resultant probabilistic models assume that the underlying random variables, which define a measurable set of outcomes, can be defined independent of the measurement context. 
The phenomenon is deemed contextual when this assumption fails.
Contextuality is an important issue in quantum physics. 
However, there has been growing speculation that it manifests outside the quantum realm with human cognition being a particularly prominent area of investigation. 
This article contributes the foundations of a probabilistic programming language that allows convenient exploration of contextuality in wide range of applications relevant to cognitive science and artificial intelligence. 
%how we did it
Using the style of syntax employed by the probabilistic programming language WebPPL, specific syntax is proposed to allow the specification of ``measurement contexts". Each such context delivers a partial model of the phenomenon based on the associated experimental condition described by the measurement context. An important construct in the syntax determines if and how these partial models can be consistently combined into a single model of the phenomenon. The associated semantics are based on hypergraphs in two ways. Firstly, if the schema of random variables of the partial models is acyclic, a hypergraph approach from relational database theory is used to compute a join tree from which the partial models can be combined to form a single joint probability distribution. Secondly, if the schema is cyclic, measurement contexts are mapped to a hypergraph where edges correspond to sets of events denoting outcomes in measurement contexts. 
Recent theoretical results from the field of quantum physics show that  contextuality can be equated  with the possibility of constructing a probabilisitic model on the resulting hypergraph. %
 %what we found
The use of hypergraphs opens the door for a theoretically succinct and efficient computational semantics sensitive to modelling both contextual and non-contextual phenomena. In addition, the hypergraph semantics allow measurement contexts to be combined in various ways. 
This aspect is  exploited to allow the modular specification of experimental designs involving both signalling and no signalling between components of the design. 
An example is provided as to how the hypergraph semantics may be applied to investigate contextuality in an information fusion setting.
%implications/significance/impact
Finally, the overarching aim of this article is to raise awareness of contextuality beyond quantum physics and to contribute formal methods to detect its presence by means of probabilistic programming language semantics.

\end{abstract} 
\end{frontmatter}
%\today{}
%\maketitle

\algnewcommand{\algorithmicgoto}{\textbf{go to}}%
\algnewcommand{\Goto}[1]{\algorithmicgoto~\ref{#1}}%
%\algnewcommand{\BlockBegin}{\algorithmicbegin}
%\algnewcommand{\BlockEnd}{\algorithmicend}
%\algblock[Block]{\BlockBegin}{\BlockEnd}
\algblock[Name]{Begin}{End}
%\newpage
\section{Introduction}
Probabilistic models are used in a broad swathe of disciplines ranging from the social and behavioural sciences, biology, the physical and computational sciences, to name but a few. 
%Large and complex probabilistic models are now being developed to cope with the demands of Big Data.
At their very core, probabilistic models are defined in terms of random variables, which range over a set of outcomes that are subject to chance.
For example, a random variable $R$ might be defined to model the performance of human memory. 
In this case, the possible outcomes might be words studied by a human subject before their memory is cued.
After cueing, the subject recalls the first word that comes to mind from the set of study words.
This outcome is recorded as a measurement.
Repeated measurements over a set of subjects allow the probability of the recall of a certain word to be empirically established.

It is important to note  from the outset that the random variable $R$ has been been devised by the modeller with a \emph{specific functional identity} in mind, namely to model the recall of a set of predefined study words.
%More generally, it is the modeller's task to define variables deemed suitable to model the problem at hand, where each variable has a specific functional identity, or purpose.
%The experiment is conducted by taking measurements whereby one of the possible outcomes is established.
When developing probabilistic models in this way, the underlying assumption is that the functional identity of a random variable is \emph{independent} of the context in which it is measured. 
For example, the purpose, or functional identity of $R$ is assumed to be the same regardless of whether the memories of human subjects are studied in a controlled laboratory, or in ``the wild", such as in a night club.
This assumption seems perfectly reasonable.
However, in quantum physics the analog of this assumption does not always hold and has become known as ``contextuality". 
More formally, the Kochen-Specker theorem \citep{kochen:specker:67} implies that quantum theory is incompatible with the assumption that measurement outcomes are determined by physical properties that are independent of the measurement context.
Placing this theorem in the context of probabilistic models: contextuality is the ``impossibility of assigning a single random variable to represent the outcomes of the same measurement procedure in different measurement conditions" \citep{barros:oas:2015}. 
%contextuality occurs when the same random variable changes its functional identity depending on the context in which it is measured \citep{dzhafarov:kujala:2014c}.

Contextuality plays a central role in the rapidly developing field of quantum information in delineating how quantum resources can transcend the bounds of classical information processing \citep{howard:2014:contextuality}. 
It also has important consequences for our understanding of the very nature of physical reality.
It is still an open question, however, if contextuality manifests outside the quantum realm. 
%\citet{Abramsky:2015} has stated that contextuality is ``pervasive". 
Some authors in the emerging field of quantum cognition have investigated whether contextuality manifests in cognitive information processing, for example, human conceptual processing \citep{Article:02:Aerts:QM,Aerts:2014,aerts:sozzo:2014,bruza:kitto:ramm:sitbon:2015,gronchi:strambini:2017} and perception \citep{atmanspacher:filk:2010,khrennikov:2014,zhang:dzhafarov:2017}. 
%In contrast,  \citet{dzhafarov:zhang:kujala:2015} conclude that behavioural and social systems are non-contextual.

It is curious that the preceding deliberations around random variables have a parallel in the field of computer programming languages.
More than five decades ago, programming languages such as FORTRAN featured variables that were global. (In early versions of FORTAN, \emph{all} variables were global.)
%For this reason they were referred to as  ``global" variables.
As programming languages developed, global variables were seen as a  potential cause of errors.  
For example, in a large program a variable $X$ can inadvertently be used for functionally \emph{different} purposes at different points in the program.
%The problem is that the programmer has inadvertently overloaded the variable; 
The error can be fixed by splitting variable $X$ into two global variables $X_1$ and $X_2$.
In this way $X_1$ can be used for one functional purpose and $X_2$ for the other, and hence there is no danger that their unique functional identities can become confounded.
However, when the program involves large numbers of global variables, keeping track of the functional identities of variables can became tedious and a  source of error. 
Such errors were considered serious and prevalent enough that following in the wake of Dijkstra's famous paper titled ``Go To statement considered harmful", \citet{wulf:shaw:1973} advocated in a similarly influential article that global variables are ``harmful" and perhaps should be abolished.
This stance was developed in relation to block structured programming languages.
A ``block", or ``scope", refers to the set of program constructs, such as variable definitions, that are only valid \emph{within} a delineated syntactic fragment of program code.
\citet{wulf:shaw:1973} argued that when a program employs a scope in which variable $X$ is defined locally, as well as a variable with the same label $X$ that is global to that scope, then $X$ becomes ``vulnerable" for erroneous overloading.
The theory of programming languages subsequently developed means so that a variable with the same label can be used in two different scopes but preserve a unique functional identity \emph{within} the given scope. 
This is not the case in state-of-the-art probabilistic modelling.
We believe that the way probabilistic models are currently developed is somewhat akin to writing FORTRAN programs from a few decades ago.
By this we mean that in the development of a probabilistic model  \emph{all} the random variables are global.
As a consequence errors can appear in the associated model should the functional identity of variables be changing because the phenomenon being modelled is contextual. 
%FIXEM this para to the conculsions
%We are of the opinion that root cause of these bugs, namely contextuality, remains undiagnosed and is possibly being masked in some cases by the introduction of parameters, or latent variables into the models. (See Briggs' insightful reflections regarding the ``mystery of parameters" \citep{briggs:2016}). 

The aim of this article to contribute the foundations of a probabilistic programming language that allows convenient exploration of contextuality in wide range of applications relevant to cognitive science and artificial intelligence. 
For example, dedicated syntax is illustrated which shows how a measurement context can be specified as a syntactic scope in a probabilistic program. In addition, random variables can be declared local to a scope to allow overloading, which is convenient for the development of models. 
Such programs are referred to a P-programs and fall within the emerging area of probabilistic programming \citep{gordon:henzinger:2014}. 

Probabilistic programming languages (PPLs) unify techniques from conventional programming such as modularity, imperative or functional specification, as well as the representation and use of uncertain knowledge. 
A variety of PPLs have been proposed (see \citet{gordon:henzinger:2014} for references), which have attracted interest from artificial intelligence, programming languages, cognitive science, and the natural language processing communities \citep{goodman:stuhlmuller:2014}.
However, unlike conventional programming languages, which are written with the intention to be executed, a core purpose of a probabilistic program is to specify a model in the form of a probability distribution.
In short, PPLs are high-level and universal languages for expressing probabilistic models.
%END FIXME
As a consequence, these languages should not be confused with probabilistic algorithms, or randomized algorithms, which employ a degree of randomness as part of their logic. 

In addition to the dedicated syntax, P-programs have a semantics based on hypergraphs which determine whether the phenomenon is contextual. 
These semantics will be based on hypergraphs in two ways: 
Firstly, a hypergraph approach from relational database theory is used to determine whether the schema of variables of the various measurement contexts is acyclic. If so, the phenomenon is being modelled is non-contextual. Secondly, if the schema is cyclic, measurement contexts are mapped to ``contextuality scenarios", which are probabilistic hypergraphs. 
Although these hypergraphs have been developed in the field of quantum physics, they provide a comprehensive general framework to determine whether the phenomenon being modelled by the P-program is contextual.
 
\section{An example P-program}

%In order to set the scene, this section summarizes some background work on P-programs \cite{bruza:abramsky:2017,bruza:2016}.
In order to convey some of the core ideas behind P-programs, Figure~\ref{pprog1} illustrates an example program where the phenomenon being modelled is two coins being tossed in four experimental conditions.
Some of these conditions induce various biases on the coins.
The syntax of the P-program is expressed in the style of a feature rich  probabilistic programming language called WebPPL\footnote{http://webppl.org/}\citep{probmods2}. 
However, the choice of the language is not significant. 
WebPPL is simply being used as an example syntactic framework. 
%Other languages could also be chosen such as the open source dynamic programming language Julia \citep{bezanson:2012}. 
\begin{figure}
%\tiny
\begin{lstlisting}[ 
           language=Python,
           showspaces=false,
           basicstyle=\ttfamily,
           numbers=left,
           numberstyle=\tiny,
           commentstyle=\color{gray}
        ]
var P1= context(){
# declare two binary random variables; 0.5 signifies a fair coin toss
     var A1 = flip(0.6)
     var B1 = flip(0.5)
# declare joint distribution across the variables A1, B1
     var p=[A1,B1]
# flip the dual coins 1000 times to form the joint distribution
     return {Infer({samples:1000},p)}
};
var P2= context(){
     var A1 = flip(0.6)
     var B2 = flip(0.3)
     var p=[A1,B2]
     return {Infer({samples:1000},p)}
};
var P3= context(){
     var A2 = flip(0.2)
     var B1 = flip(0.5)
     var p=[A2,B1]
     return {Infer({samples:1000},p)}
};
var P4= context(){
     var A3 = flip(0.5)
     var B2 = flip(0.3)
     var p=[A3,B2]
     return {Infer({samples:1000},p)}
};
# return a single model
return {model(P1,P2,P3,P4)}    
\end{lstlisting}
\vspace{-0.5cm}
\caption{Example P-program in the style of WebPPL.} 
\label{pprog1}
\end{figure}

In P-programs, syntactic scopes are delineated by the reserved word \texttt{context}.
Each scope specifies an experimental condition, or ``measurement context" under which a phenomenon is being examined.
The example P-program defines four such contexts labelled \texttt{P1}, \texttt{P2}, \texttt{P3} and \texttt{P4}.
Consider context \texttt{P1} which declares two coins as dichotomous random variables $A1$ and $B1$ which are local to this scope. 
The syntax \texttt{flip(0.5)} denotes a fair coin; any value other than 0.5 defines a biased coin.
Declaring variables local to the scope syntactically expresses the assumption that the variables retain a \emph{unique} functional identity within the scope.
The random variable declarations within a scope define a set of events which correspond to outcomes which can be observed in relation to the phenomenon being examined in the given measurement context.
For example, $A1=1$ signifies that coin A1 has been observed as being a head after flipping.
Joint event spaces are defined by the syntax \texttt{p[A1,B1]} which becomes a joint probability distribution by the syntax \texttt{Infer({samples:1000},p)}.
In this case two coins have been flipped 1000 times to prime the probabilities associated with each of the four mutually exclusive joint events in the event space. 
%These are used to derive a joint probability distribution which is returned by the context.
The resulting distribution represents the model of the phenomenon in that measurement context which is returned from the scope as a partial model.
The other measurement contexts \texttt{P2}, \texttt{P3} and \texttt{P4} are similarly defined resulting in the four distributions depicted in Figure \ref{fig:acyclic}.
The structure of each distribution will be referred to as a probabilistic table, or p-table for short, as these are a natural probabilistic extension to the tables defined in relational databases \citep{bruza:abramsky:2017}.
\begin{figure}
\centering
P1:
\begin{filecontents*}{P11.csv}
A1,B1,p
1,1,$p_1$
1,0,$p_2$
0,1,$p_3$
0,0,$p_4$
\end{filecontents*}
\csvautotabular{P11.csv}
%\hfill[1cm]
P2:
\begin{filecontents*}{P12.csv}
A1,B2,p
1,1,$q_1$
1,0,$q_2$
0,1,$q_3$
0,0,$q_4$
\end{filecontents*}
\csvautotabular{P12.csv}\\~\\
P3:
\begin{filecontents*}{P21.csv}
A2,B1,p
1,1,$r_1$
1,0,$r_2$
0,1,$r_3$
0,0,$r_4$
\end{filecontents*}
\csvautotabular{P21.csv}
%\hfill[1cm]
P4:
\begin{filecontents*}{P32.csv}
A3,B2,p
1,1,$s_1$
1,0,$s_2$
0,1,$s_3$
0,0,$s_4$
\end{filecontents*}
\csvautotabular{P32.csv}
\caption{Four p-tables returned by the respective contexts P1, P2, P3, P4 from the P-program of Figure \ref{pprog1}. The values in the column labelled ``p" denote probabilities and sum to unity in each table.}
\label{fig:acyclic}
\end{figure}

Modelling practice is usually governed by  the norm that it is desirable to construct a single model of the phenomenon being studied.
Dedicated syntax, e.g., \texttt{model(P1,P2,P3,P4)} allows partial models from the four measurement contexts to be combined to form a single distribution, such that each distribution corresponding to partial models can be recovered from this single distribution by appropriately marginalizing it \citep{bruza:abramsky:2017,bruza:2016}. 
It turns out that this is not always possible to construct such a single model. 
As we shall see below, when this happens the phenomenon being modelled turns out to be ``contextual".
%Conversely, when single model can be constructed corresponds to all random variables retaining a single functional identity throughout the program. Said otheriwse, their functional identities are  \emph{independent} of measurement context. 
\citet{Abramsky:2015} discovered that this formulation of contextuality is equivalent to the universal relation problem in database theory.
This problem involves determining whether the relations in a relational database be joined together (using the natural join operator) to form a single ``universal" relation such that each constituent relation be can be recovered from the universal relation by means of projection.
%It is well known from database theory that it is not always possible to construct the universal relation.

Relational database theory tells us that a key consideration in this problem turns out to be whether the database schema comprising constituent relations (p-tables) is cyclic or acyclic.
A database schema is deemed ``acyclic" iff the hypergraph $H(N,E)$ can be reduced into an empty graph using the Graham procedure \citep{gyssens:paradaens:84}.
The set $N$ denotes the vertices in the graph and $E$ the set of edges.

A hypergraph differs from a normal graph in that an edge can connect more than two vertices. For this reason, such edges are termed ``hyperedges". 
For example, the database schema corresponding to the P-program in Figure \ref{pprog1} is depicted in Figure \ref{fig:acyclic}.
In our case, the nodes $N$ of the hypergraph are the the individual variables in the headers of the p-tables and the edges correspond to the sets of variables in these headers, i.e., there will be one edge corresponding to each constituent p-table, where the edge is the set of variables defining the header of that p-table. 
Therefore, $N = \{A_1,A_2,A_3,B_1,B_2\}$ and $E = \{\{A_1,B_1\},\{{A_1,B_2\}},\{A_2,B_1\},\{A_3,B_2\}\}$.
(As the headers of p-tables only contain two variables, $H$ is in this case a standard graph.)

The Graham procedure is applied to the hypergraph $H$ until no further action is possible:
\begin{itemize}
\item delete every edge that is properly contained in another one;
\item delete every node that is only contained in one edge.
\end{itemize}
The following details the steps of the Graham procedure when applied to the example:
\begin{enumerate}
\item $\{A_1,B_1\},\{A_1,B_2\},\{A_2,B_1\},\{A_3,B_2\}$
\item $\{A_1,B_1\},\{A_1,B_2\},\{A_2,B_1\},\{B_2\}$
\item $\{A_1,B_1\},\{A_1,B_2\},\{A_2,B_1\}$
\item $\{A_1,B_1\},\{A_1,B_2\},\{B_1\}$
\item $\{A_1,B_1\},\{A_1,B_2\}$
\item $\{A_1,B_1\},\{A_1\}$
\item $\{A_1,B_1\}$
\item $\{A_1\}$
\item $\emptyset$
\end{enumerate}
In this case the Graham procedure results in an empty hypergraph, so the schema is deemed ``acyclic".
%Note, however, that the set of edges corresponding to P-program 2 in figure \ref{P-programEPR} is
%\[E = \{\{A_1,B_1\},\{{A_1,B_2\}},\{A_2,B_2\},\{A_2,B_1\}\}\]
%and the hypergraph cannot be reduced to an empty set of edges, so the schema of this P-program is termed ``cyclic".

There are a number of theoretical results in relational database theory which make acyclic hypergraphs significant with regard to providing the semantics of joining partial models into a single model.
%\citet{wong:1997} detailed the relationship between a Markov network and a relational database model.
%This connection allows a Markov distribution to be constructed by joining component distributions together.
% in much the same way as tables in a relational database can be joined to form a single table.
\citet{wong:1997} formalizes the relationship between Markov distributions and relational database theory by means of a generalized acyclic join dependency (GAJD).
The key idea behind this relationship is the equivalence between probabilistic conditional independence and a generalized form of multivalued dependency, the latter being a functional constraint imposed between two sets of attributes in the database schema.
It turns out that a joint distribution factorized on an acyclic hypergraph is equivalent to a GAJD \citep{wong:2001}.

For example, consider once again the acyclic schema in Figure \ref{fig:acyclic}.
There are four p-tables, $P=\{P_1,P_2,P_3,P_4\}$. 
As the hypergraph is acyclic, there is necessarily a so called join tree construction, denoted $\otimes\{S_1,\ldots,S_n\}$, that satisfies the GAJD. 
In the tree construction, each $S_i, 1\leq i \leq n$ denotes a unique p-table in the set $P$.
The practical consequence of this is that there is a join expression of the form: $(((S_1 \otimes S_2) \otimes S_3) \otimes S_4)$ where the sequence $S_1,\ldots, S_4$ is a tree construction ordering derived from the acyclic hypergraph.  
If the hypergraph constructed from the  schema comprising  $n$ p-tables $\{P_1,P_2,P_3,\ldots P_n\}$ is acyclic, then a generalized join expression 
$(\ldots(S_1 \otimes S_2) \otimes S_3) \ldots \otimes S_n)$ exists which joins the p-tables into a  single probability distribution $P$ such that each $P_i, 1\leq i \leq n$ is a marginal distribution of $P$.

\begin{figure}[h]
\centering
\includegraphics[width=7cm]{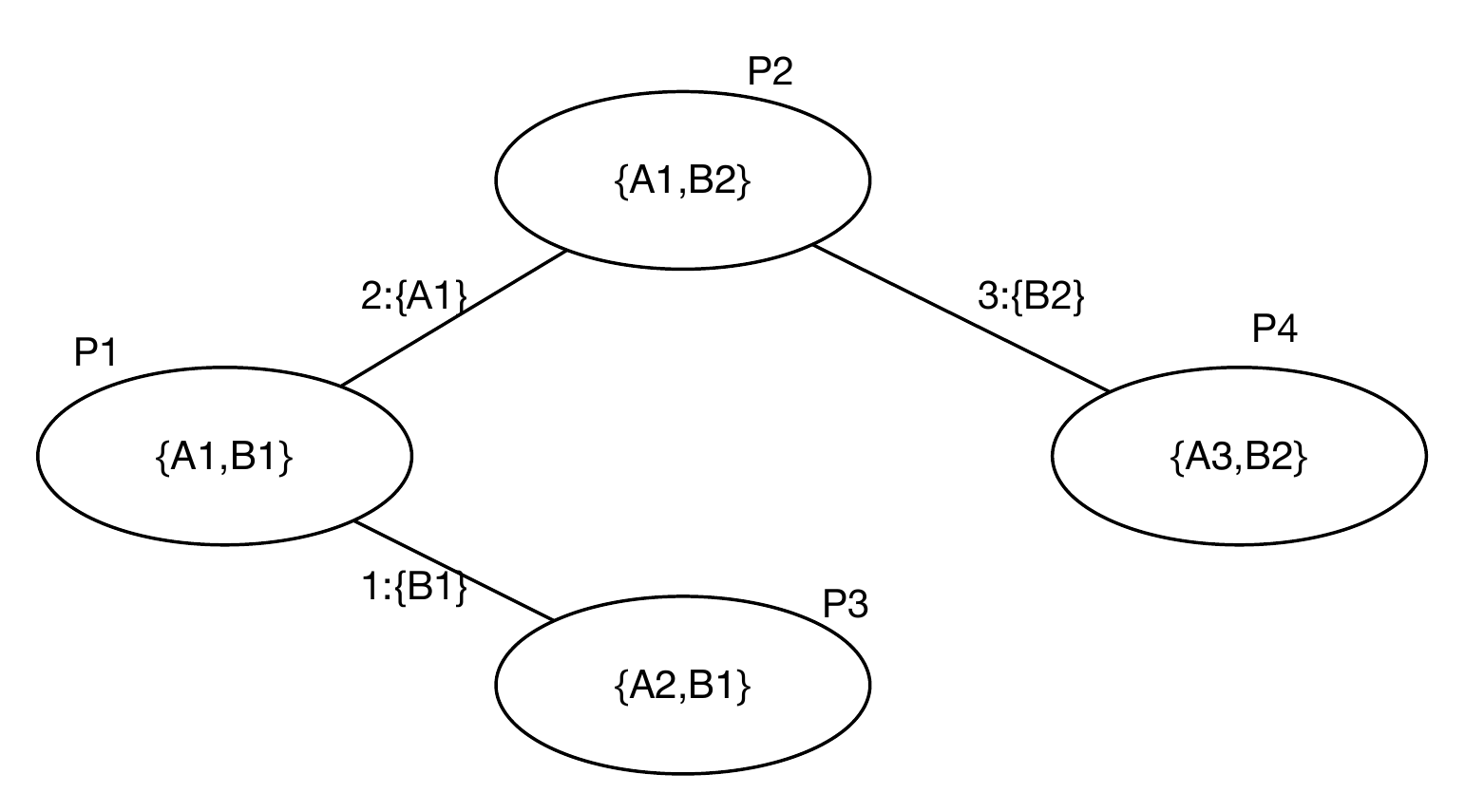}
\caption{Join tree of the p-tables $P_1,P_2,P_3,P_4$ to be joined.}\label{fig:jointree}
\end{figure}
In order to gain some intuition about how this plays out in practice, the acyclic database schema depicted in Figure \ref{fig:acyclic} results in the join tree depicted in Figure \ref{fig:jointree}.
The nodes depict the variables in the respective p-tables and the edges represent the overlap between the sets of variables in the respective headers.
The numbers on the edges denote the ordering used to produce the join expression: $(((P_3 \otimes P_1) \otimes P_2) \otimes P_4)$.
Under the assumption that the probability distributions represented in the nodes have identical distributions when marginalized by the variable associated with the edge, we can see how the hypertree produces a Markov network which, in turn,  specifies the probabilistic join of the constituent p-tables \citep{liu:yue:li:2011}:
\begin{align}
P(A_1,A_2,A_3,B_1,B_2) &= \frac{P(A_2,B_1)P(A_1,B_1)P(A_1,B_2)P(A_3,B_2)}{P(B_1)P(A_1)P(B_2)} \label{eqn:markov}
\end{align}
Observe how the structure of the equation mirrors the graph in Figure \ref{fig:jointree} where the numerator corresponds to the nodes of the join tree and the denominator corresponds to terms which normalize the probabilities. 
In addition, this expression reflects conditional independence assumptions implied by the join tree, namely $A_1$ and $A_2$ are conditionally independent of $B_1$, $B_1$ and $B_2$ are conditionally independent of $A_1$, and $A_1$ and $A_3$ are conditionally independent of $B_2$. 
%The key idea behind this connection was the discovery of the equivalence between probabilistic conditional independence and a generalized form of multivalued dependency (GEMVD).
%Multivalued dependencies are a particular form of mathematical function which are used define constraints on the data stored within the database.

Let us summarize the situation so far and reflect on the issue of contextuality. 
A P-program comprises a number of scopes where each scope corresponds to a measurement context. 
A scope returns a probability distribution in the form of a p-table, which can be considered a partial model of the phenomenon. 
A reasonable goal is to combine these distributions into a single distribution so a single model of the phenomenon is produced.
When the schema of the constituent p-tables is acyclic and the marginal distributions of the set of intersecting variables  are constant, then a straightforward extension of relational database theory can be used to produce the required single model (as has been shown in more detail in \citep{bruza:2016}).
The fact that it is possible to construct a single model means that the random variables in the P-program have a functional identity that is \emph{independent} of the measurement contexts. 
%In other words, they preserve a single functional identity across measurement contexts in relation to modelling the phenomenon in question. 
The phenomenon is therefore \emph{non-contextual}.

Much of the research on contextuality corresponds to when the schema of the p-tables is cyclic \citep{zhang:dzhafarov:2017}.
In order to explore such cases, we will continue to use hypergraphs, but instead on defining the graph structure at the level of the schema as was illustrated in Graham procedure, the structure of the hypergraph will be defined in terms of the underlying events in the measurement contexts defined in the P-program.
In database terms this equates to defining the hypergraph structure in terms of the data in the respective p-tables.

\section{Probabilistic models and Hypergraph semantics}

In the following, we draw from a comprehensive theoretical investigation using hypergraphs to model contextuality in quantum physics \citep{Acin:2015}. 
The driving motivation is to leverage these theoretical results to provide the semantics of P-programs when the schema of the p-tables to be joined is cyclic. 
How these semantics are expressed relates to how the syntax has been specified, which in turn relates to the experimental design that the modeller has in mind. 
The basic building block of these semantics is a ``contextuality scenario". 
\begin{definition}{(Contextuality Scenario)} (Definition 2.2.1  \citep{Acin:2015})
A \emph{contextuality scenario} is a hypergraph $X=(V,E)$ such that:
\begin{itemize}
\item $v \in V$ denotes an event which can occur in a measurement context;
\item $e \in E$ is the set of all possible events in a measurement context.
\end{itemize}
\end{definition}
The set of hyperedges $E$ are determined by both the measurement contexts as well as the measurement protocol. 
Each measurement context is represented by an edge in the hypergraph $X$. 
The basic idea is that each syntactic scope in a P-program will lead to a hyperedge, where the events are a complete set of outcomes in the given measurement context specified in the associated scope. 
Additional hyperedges are a consequence of the constraints inherent in the measurement protocol that is applied.
The examples to follow aim to make this clear.

%We shall see subsequently that the events correspond to the rows of p-tables returned by a context in a P-program.

In some cases, hyperedges will have a non-trivial intersection:  
If $v \in e_1$ and $v \in e_2$, then this represents the idea that the two different measurement outcomes corresponding to  $v$ should be thought of as equivalent as will be detailed below by means of an order effects experiment.
%If $v \in e_1$ and $v \in e_2$ means that the event $v$ is \emph{equivalent} across measurement contexts $e_1$ and $e_2$.
%By way of illustration, the field of psychology sometimes conducts order effects experiments. 

Order effects experiments involve two measurement contexts each involving two dichotomous variables $A$ and $B$ which represent answers to yes/no questions $Q_A$ and $Q_B$.
In one measurement context, the question $Q_A$ is served before question $Q_B$ and in the second measurement context the reverse order is served, namely $Q_B$ then $Q_A$. 
Order effects occur when the answer to the first question influences the answer to the second.
%Order effects experiments adhere to a ``between subjects" design as it important that a given subject be subjected to only one of the experimental contexts.
%Between subject designs  can be directly translated into syntactic scopes as depicted in the order effects P-program shown in figure \ref{fig:order-effects}.
These two measurements contexts are syntactically specified by the scopes \texttt{P1} and \texttt{P2} shown in Figure \ref{pprog:order-effects}.
\begin{figure}
\small
\begin{lstlisting}[ 
           language=Python,
           showspaces=false,
           basicstyle=\ttfamily,
           numbers=left,
           numberstyle=\tiny,
           commentstyle=\color{gray}
        ]
var P1 = context() {
     var A = flip(0.7)
     var B = A ? flip(0.8): flip(0.1)
     var p=[A,B]
     return {Infer({samples:1000},p}
};
var P2 = context() {
     var B = flip(0.4)
     var A = B ? flip(0.4): flip(0.6)
     var p=[B,A]
     return {Infer({samples:1000},p}
}; 
return {model(P1,P2)}
\end{lstlisting}
\vspace{-0.5cm}
\caption{Example order effects P-program.} 
\label{pprog:order-effects}
\end{figure}

In this P-program, syntax of the form 
\texttt{var B = A ? flip(0.8): flip(0.1)} models the influence of the answer of $Q_A$ on $Q_B$ via a pair of biased coins. 
In this case, if $Q_A=y$, then the response to $Q_B$ is determined by flipping an 80\% biased coin. 
Conversely, if $Q_A=n$, then the response to $Q_B$ is determined by flipping a 10\% biased coin
(The choices of such biases are determined by the modeller).
% insert Belen sainz
It should be carefully noted that the measurement contexts in the order effects program do not reflect
the usual understanding of measurement context employed in experiments analyzing contextuality in quantum physics. In these experiments, a measurement context comprises observables that are jointly measurable, so the order in which the observables within a given context are measured will not affect the associated statistics. 

\begin{figure}[h!]
\centering
\includegraphics[width=5cm]{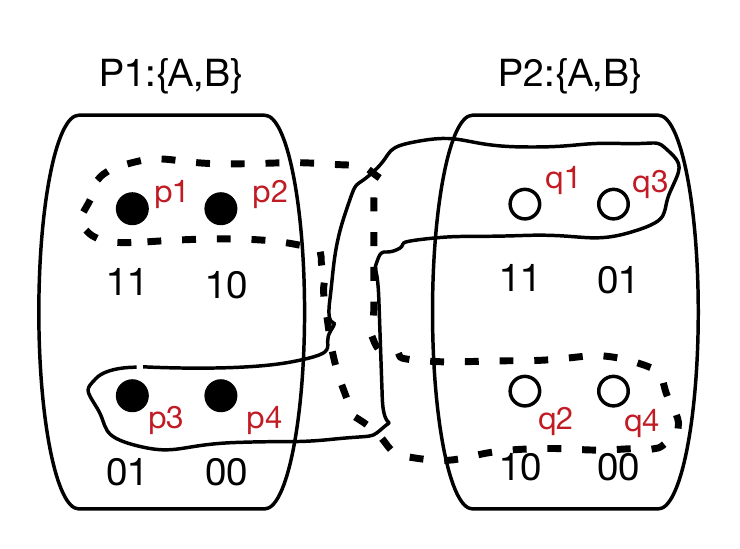}
\caption{Contextuality scenario corresponding to P-program depicted in Figure \ref{pprog:order-effects}.  In total, the hypergraph has 4 edges of four vertices each.}
\label{fig:hyper-order}
\end{figure}

We will now use this simple example to illustrate the associated contextuality scenario which is shown in Figure \ref{fig:hyper-order}. Firstly, the set of $V$ of events (measurement outcomes) comprises all possible combinations of yes/no answers to the questions $Q_A $ and $Q_B$, namely $V = \{A=1 \wedge B=1, A=1 \wedge B=0,A=0 \wedge B=1,A=0 \wedge B=0\}$, where 1 denotes `yes' and 0 denotes `no'. 
In this figure, the two rounded rectangles represent the events within the two measurement contexts specified by the syntactic scopes P1 and P2. 
For example, in the rectangle labeled P1, ``11" is shorthand for the event $A=1 \wedge B=1$ etc.
Observe that the corresponding hyperedges (rounded rectangles) contain an exhaustive, mutually exclusive set of events.
In addition, the two spanning hyperedges going across these rectangles similarly comprise a set of exhaustive, mutually exclusive set of events.
These spanning edges help illustrate events that are considered to be equivalent.

Firstly, it is reasonable to assume answering yes (or no) to both questions in either measurement context represents equivalent events.
Therefore, the events labelled $p_1$ and $p_4$ can respectively be assumed equivalent to $q_1$ and $q_4$.
It becomes a little more subtle when the polarity of the answers differ.
For example, the event labelled  $p_3$  represents the event $A=0 \wedge B=1$, remembering that question $Q_A$ was asked before question $Q_B$ in this context.
The equivalent event in hyperedge  P2 is labelled $q_2$, which corresponds  the event $B=1 \wedge A=0$, where question $B$ is asked before question $A$. 
As conjunction is commutative, it is reasonable to view these two converse events as equivalent.
In summary, if $p_3$ is equivalent to $q_2$ and $p_4$ is equivalent to $q_4$ then the hyperedge 
$\{p_1,p_2,q_2,q_4\}$ (the dashed hyperedge in Figure \ref{fig:hyper-order}) can be established, in addition to the hyperedge $\{p_1,p_2,p_3,p_4\}$.
%Note how defining equivalent events in this way leads to two hyperedges with a non-empty overlap by virtue of the events labelled $p_1$ and $p_2$.

Let us now return to the issue of contextuality. 
A \emph{probabilistic model} corresponding to a contextuality scenario $X$ is the mapping of measurement outcomes to a probability $p: V \rightarrow [0,1]$. 
\citet{henson:2015} point out that 
\begin{quote}
``By defining probabilistic models in this way [rather than by a function $p_e(V)$ depending on the measurement $e$ performed], we are assuming that in the set of experimental protocols that we are interested in, the probability for a given outcome is independent of the measurement that is performed".
\end{quote}
Observe carefully that by defining probabilistic models in this way formalizes the assumption mentioned in the introduction, namely that random variables are \emph{independent} of measurement context and thus have a single functional identity.  
Without a single functional identity it is \emph{impossible} to assign a random variable to represent the outcomes of the same measurement protocol in different measurement contexts.

It is a requirement that the mapping adheres to the expected normalization condition: $\forall_{e \in E}:\sum_{v \in e}p(v)=1$. 
By way of illustration, consider once again Figure \ref{fig:hyper-order}. 
This contextuality scenario has four edges. The normalization condition enforces the following constraints:
\begin{align}
p_1 + p_2 + p_3 + p_4 = 1 \label{eq1}\\
q_1 + q_2 + q_3 + q_4 = 1 \label{eq2}\\
p_1 + p_2 + q_2 + q_4 = 1  \label{eq3}\\
p_3 + p_4 + q_1 + q_3 = 1  \label{eq4}
\end{align}
where $p_i, 1 \leq i \leq 4$ and $q_j, 1 \leq j \leq 4$ denote the probabilities of outcomes in the four hyperedges. A definition of contextuality can now be presented.
\begin{definition}[(Probabilistic) contextuality] 
Let $X=(V,E)$ be a contextuality scenario. Let $\pX{X}$ denote the set of probabilistic models on $X$. $X$ is deemed ``contextual" if $\pX{X} = \emptyset$.
\label{def:contextuality}
\end{definition}
In other words, the impossibility of a probabilistic model signifies that the phenomenon being modelled is contextual.
(The label ``probabilistic" mirrors an analogous definition of contextuality based on sheaf theory \citep{abramsky:barbosa:mansfiled:2016}).
%For example, consider once again the contextuality scenario shown in figure \ref{fig:hyper-order}.
%Recall that this contextuality scenario expresses the hypergraph semantics associated with the order effects P-program depicted in figure \ref{pprog:order-effects}. 

Let us now examine the possibility of  a probabilistic model on the order effects contextuality scenario (Figure \ref{fig:hyper-order}). 
%Equations \eqref{eq1} and \eqref{eq3} imply that $p_1 +p_3 = q_2+q_4$. 
Equations \eqref{eq1} and \eqref{eq4} imply that $p_1 +p_2 = q_1+q_3$.
Now, $p_1 + p_2$ are repectively associated with the outcomes $A=1 \wedge B=1$ and $A=1 \wedge B=0$. 
In other words, $p_1 + p_2$ denotes the marginal probability  $p(A)$ in measurement context P1.
By a similar argument, $q_1+q_3$ denotes $p(B=1 \wedge A=1) + p(B=0 \wedge A=1)$ which is written this way to emphasize that question $Q_B$ is asked first in measurement context P2. 
This also equates to the marginal probability  $p(A)$.
In other words, the constraints imposed by normalization conditions in the hyperedges imply that the marginal probability $p(A)$ must be the same across both measurement contexts P1 and P2.

This conclusion makes sense when considered in relation to the definition of contextuality:
% In the order effects contextuality scenario there are only two variables $A$ and $B$.
The only way that a function $p: V \rightarrow [0,1]$ can be defined is if the marginal probabilities of the variables $A$ and $B$ are the same in \emph{both} measurement contexts P1 and P2. 
%If not, the functional identities of the associated random variables are dependent on the measurement context. 
If not, then this means that variable $A$ has a different functional identity when question $Q_A$ is asked first (in measurement context P1) as opposed to when it is asked second (in measurement context P2).
%In that case, there is an order effect between the questions as there is an influence transpiring between them and the analysis just presented deems such a situation as strongly contextual. 

In summary,  the semantics of a P-program is represented by a contextuality scenario, which has the form of a hypergraph. 
Contextuality equates to the impossibility of a probabilistic model over the hypergraph.
This impossibility is where contextuality meets probabilistic models. 

\section{Syntax and semantics of combining contextual scenarios according to experimental design}

Different fields employ various experimental designs when studying a phenomenon of interest.
For example, in psychology a ``between subjects" experimental design means a given participant should only be subjected to one measurement context.
In quantum physics, however, some experiments involve measurement contexts which are enacted simultaneously with the requirement that observations made in each context are local to that context and don't influence other measurement contexts. 
This constraint is often referred to as the  ``no signalling" condition. 
%So called Bell experiments\footnote{The original design of these experiments was proposed by the physicist John Bell} involving entangled photons are a prominent example of where the no signalling condition is central to the validity of the experiment. 

\begin{figure}[h]
\centering
 \includegraphics[width=8cm]{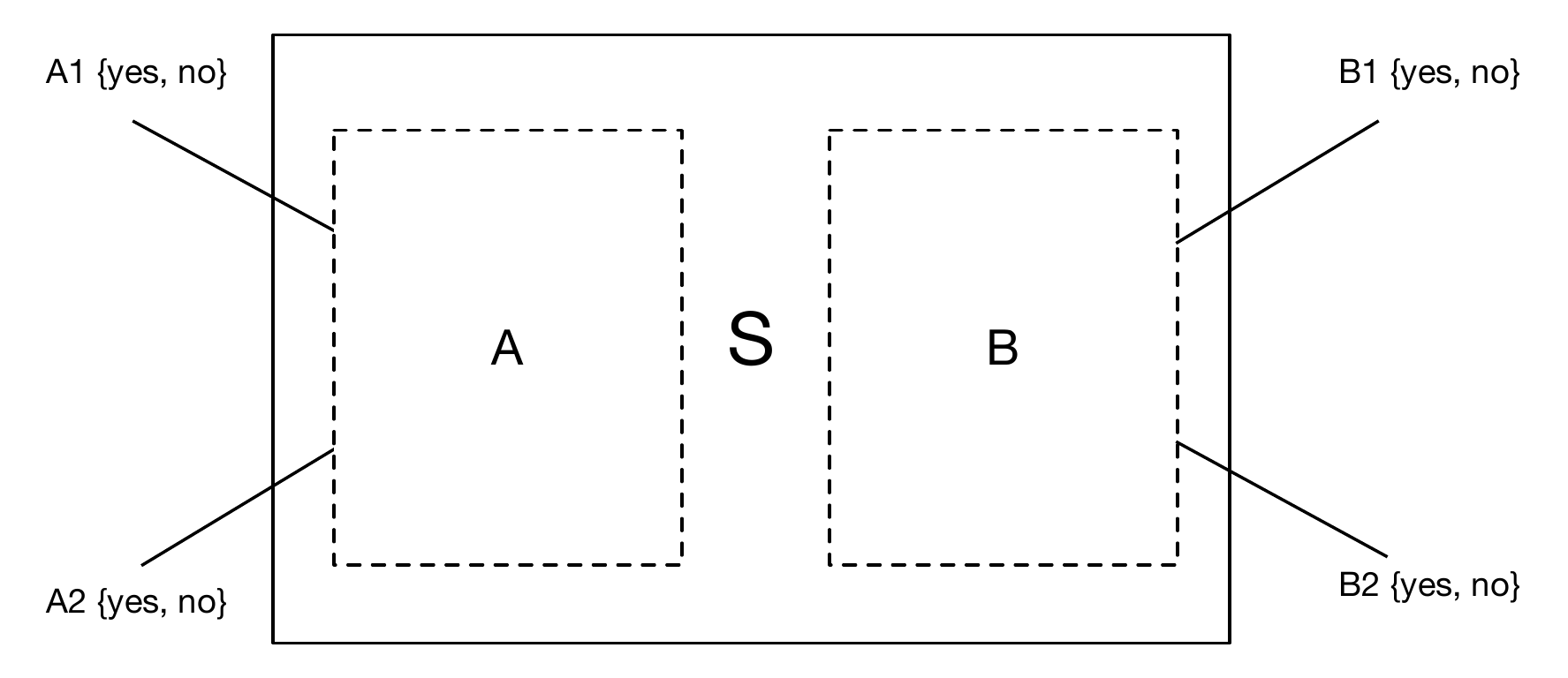}
\caption{A potentially compositional system $S$, consisting of two assumed components $A$ and $B$. $S$ can perhaps be understood in terms of a mutually exclusive choice of experiments performed upon those components, one represented by the random variables $A1,A2$ (pertaining to an interaction between the experimenter and component $A$), and the other by $B1,B2$ (pertaining to an interaction between the experimenter and component $B$). Each of these experiments can return a value of `yes' or `no'.}\label{fig:system}
\end{figure}

One of the advantages of using a programming approach to develop probabilistic models is that experimental designs can be syntactically specified in a modular way. 
In this way, a wide variety of experimental designs across fields can potentially be catered for. 
For example, consider the situation where an experimenter wishes to determine whether a system  $S$ can  validly be modelled compositionally in terms of two component subsystems $A$ and $B$ as shown in Figure~\ref{fig:system}.
%A `black box' is depicted, with two proposed components, $A$ and $B$, inside it. 

Two different experiments can be carried out upon each of the two presumed components, which will answer a set of `questions' with binary outcomes, leading to four measurement contexts. 
For example, one experimental context would be to ask $A1$ of component $A$ and $B1$ of component $B$.

This abstract experimental design has be instantiated in a number of ways.
For example, in quantum physics it has been employed to determine whether system $S$ comprising photons\footnote{Here a bipartite system of photons is being introduced. Whenever such systems of photons are mentioned throughout this article, in principle, any bipartite or multipartite quantum system would do, even fermions}.
 $A$ and $B$ is entangled.
In addition, it has been employed in cognitive psychology to test for contextuality in human cognition \citep{Aerts:2014,aerts:sozzo:2014,bruza:kitto:ramm:sitbon:2015,dzhafarov:zhang:kujala:2015,gronchi:strambini:2017}.
For example, \citet{bruza:kitto:ramm:sitbon:2015} describe an experiment to determine whether novel conceptual combinations such as BOXER BAT adhere to the principle of semantic compositionality \citep{pelletier:1994}. 
Semantic compositionality entails that the meaning of BOXER BAT is some function of the meaning of the component concepts BOXER and BAT.
In this case, component $A$ corresponds to the concept BOXER and component $B$ corresponds to the concept BAT. 
Each of these concepts happen to be bi-ambiguous, for example, BOXER can be interpreted in a sport sense or an animal sense (a breed of dog). 
Similarly, the concept BAT can be interpreted in either of these senses.
Interpretation of concepts can be manipulated by priming words which correspond to the `questions' asked of the component concepts.
For example, one experimental context would be to ask a set of human subjects to return an interpretation of BOXER BAT after being shown the priming  words \emph{fighter} ($A1$) and \emph{vampire} ($B1$). 
Note how $A1$ is designed to prime the sport sense of BOXER and $B2$ to prime the animal sense of BAT.
An interpretation given in this context might be ``an angry furry black animal with boxing gloves on".
It is important to note that the interpretation is probabilistic, namely the priming word influences an interpretation of the concept but does not determine it.

How can system $S$ depicted in Figure \ref{fig:system} be modelled as a P-program? And, how can the semantics of the P-program determine whether $S$ is contextual?
One way to think about system $S$ is that it is equivalent to a set of biased coins $A$ and $B$, where the bias is local to a given measurement context.
Figure \ref{pprog:EPR} depicts a P-program that follows this line of thinking.
%\newpage
\begin{figure}
\small
\begin{lstlisting}[ 
           language=Python,
           showspaces=false,
           basicstyle=\ttfamily,
           numbers=left,
           numberstyle=\tiny,
           commentstyle=\color{gray}
        ]
# define the components of the experiment
def A = component(A1,A2)
def B = component(B1,B2)        

var P1= context(){
# declare two binary random variables; 0.5 signifies a fair coin toss
     var A1 = flip(0.6)
     var B1 = flip(0.5)
# declare joint distribution across the variables A1, B1
     var p=[A1,B1]
# flip the dual coins 1000 times to form the joint distribution
     return {Infer({samples:1000},p)}
};
var P2= context(){
     var A1 = flip(0.4)
     var B2 = flip(0.7)
     var p=[A1,B2]
     return {Infer({samples:1000},p)}
};
var P3= context(){
     var A2 = flip(0.2)
     var B1 = flip(0.7)
     var p=[A2,B1]
     return {Infer({samples:1000},p)}
};
var P4= context(){
     var A2 = flip(0.4)
     var B2 = flip(0.5)
     var p=[A2,B2]
     return {Infer({samples:1000},p)}
};
# return a single model
return {model({design: `no-signal',P1,P2,P3,P4})}    
\end{lstlisting}
\vspace{-0.5cm}
\caption{Example P-program ``Bell scenario" which models system S depicted in Figure \ref{fig:system}.} 
\label{pprog:EPR}
\end{figure}

The P-program will be referred to as ``Bell scenario" as it programmatically specifies the design of experiments in quantum physics inspired by the physicist John Bell \citep{clauser:horne:74}.
Such experiments involve a system of two space-like separated photons. 
%In this way measuring the polarization of one photon should not affect the other, i.e., the ``no signalling" condition holds. 
\subsection{Bell contextuality scenario with no-signalling}
The Bell scenario program follows the design depicted in Figure \ref{fig:system} by first defining the components $A$ and $B$ together with the associated variables.
Thereafter, the program features the four measurement associated contexts \texttt{P1}, \texttt{P2}, \texttt{P3} and \texttt{P4}. 
Finally, the line \texttt{model({design: 'no-signal',P1,P2,P3,P4})} specifies that the measurement contexts are to be  combined according to a ``no signalling" condition.
Formal details of this condition will follow, but essentially it imposes a constraint that measurements made on one component do not affect outcomes observed in relation to the other component.
This could be because the components have sufficient spatial separation in a physics experiment, or alternatively, in a psychology experiment the cognitive phenomena represented by components $A$ and $B$ of system $S$ are independent cognitive functions. 

\begin{figure}[h]
\centering
 \includegraphics[width=6cm]{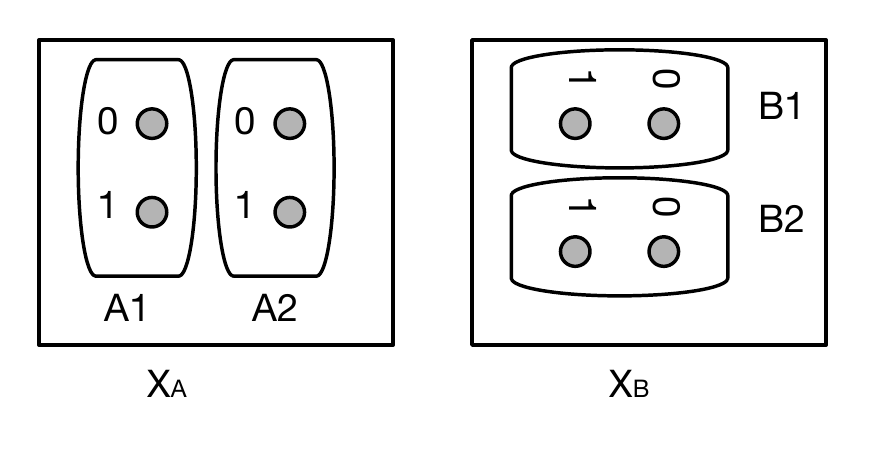}
\caption{Contextuality scenarios corresponding to the components A and B of Figure \ref{fig:system}.}\label{fig:XaXb}
\end{figure}

The question now to be addressed is how the hypergraph semantics are to be formulated. 
\citet{Acin:2015} provides the general semantics of the Bell scenarios  by means of multipartite composition of contextuality scenarios.
As these semantics are compositional, it opens the door to map syntactically specified components in a P-program to  contextuality scenarios and then to exploit the composition to provide the semantics of the program as a whole.
Consider the Bell scenario program depicted in Figure \ref{pprog:EPR}.
The syntactically defined components  $A$ and $B$ are modelled as a contextuality scenarios $X_{A}$ and $X_{B}$ respectively.
The corresponding hypergraphs are depicted in Figure \ref{fig:XaXb}. 

Note how the variable definitions associated with the component map to an edge in a hypergraphs.
For example, the syntax \texttt{def A = component(A1,A2)} corresponds to the two edges labelled $A1$ and $A2$ on the left hand side of Figure \ref{fig:XaXb}.

\begin{figure}[h]
\centering
\includegraphics[width=7cm]{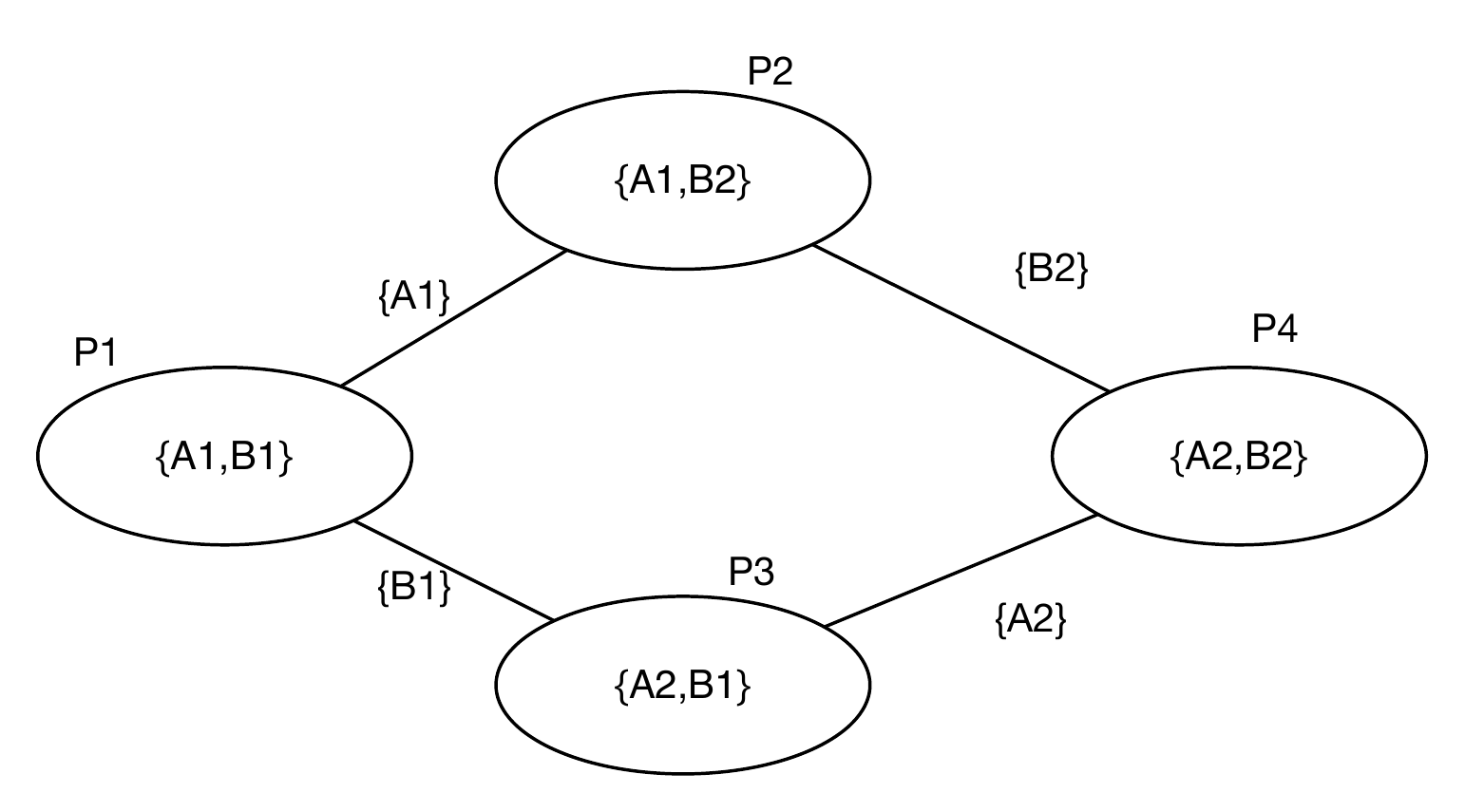}
\caption{Cyclic schema of the p-tables P1, P2, P3, and P4.}\label{fig:cyclic}
\end{figure}

\begin{figure}[h]
\centering
 \includegraphics[width=5cm]{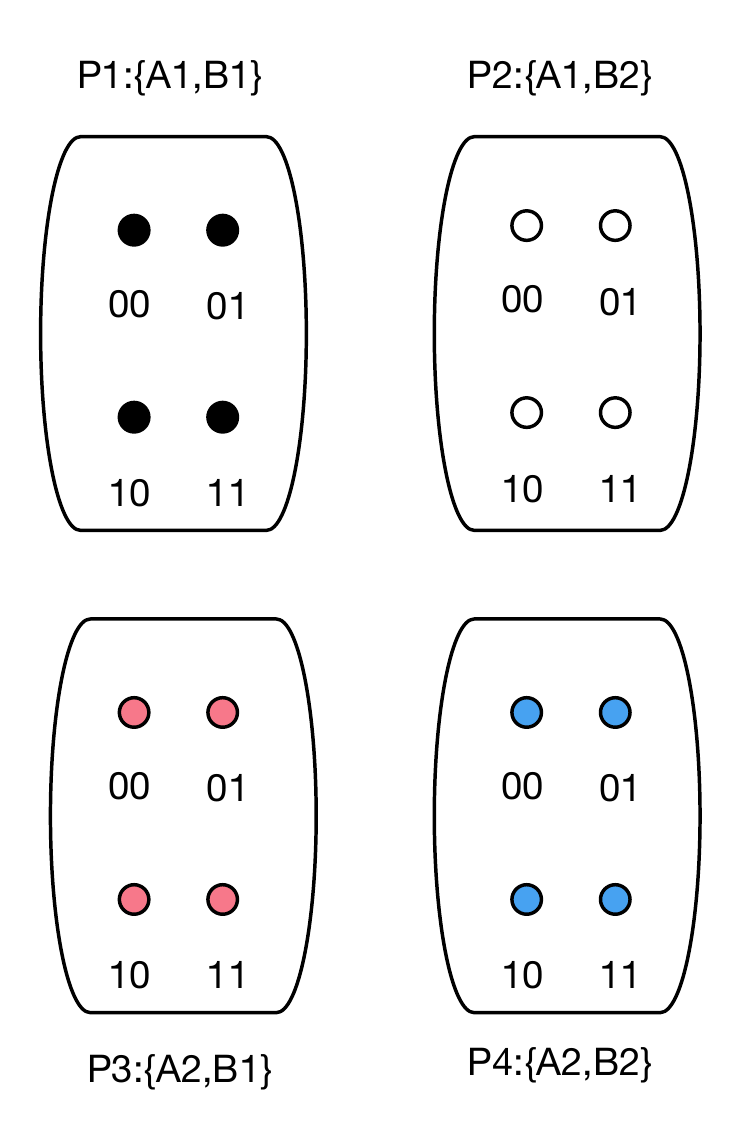}
\caption{Contextuality scenario corresponding to the direct product $X_{AB}=X_A \times X_B$  \label{fig:product}}
\end{figure}

The question now is how the compose the contextuality scenarios $X_{A}$ and $X_{B}$  into a single contextuality scenario $X_{AB}$, which will express the semantics of the Bell scenario P-program.
The most basic form of composition is by means of  a direct product of the respective hypergraphs. 
The direct product is a contextual scenario $X_{AB} = X_A \times X_B$ such that $V(X_A \times X_B) = V(X_A) \times V(X_B)$ and $E(X_A \times X_B) = E(X_A) \times E(X_B)$. (See Definition 3.1.1 in \citep{Acin:2015}.)
The hypergraph of the product is shown in Figure \ref{fig:product}.
Observe how each syntactic context \texttt{P1}, \texttt{P2}, \texttt{P3} and \texttt{P4} specified in the Bell scenario P-program corresponds to an edge in the hypergraph.
In addition, note the structural correspondence of the hypergraph in Figure \ref{fig:product} with the  cyclic database schema depicted in Figure \ref{fig:cyclic}.

Note that the events in Figure \ref{fig:product} are denoted as various coloured dots with each such dot corresponding directly to a row of a p-table within the cyclic schema. 

\begin{figure}[h]
\centering
\includegraphics[width=9cm]{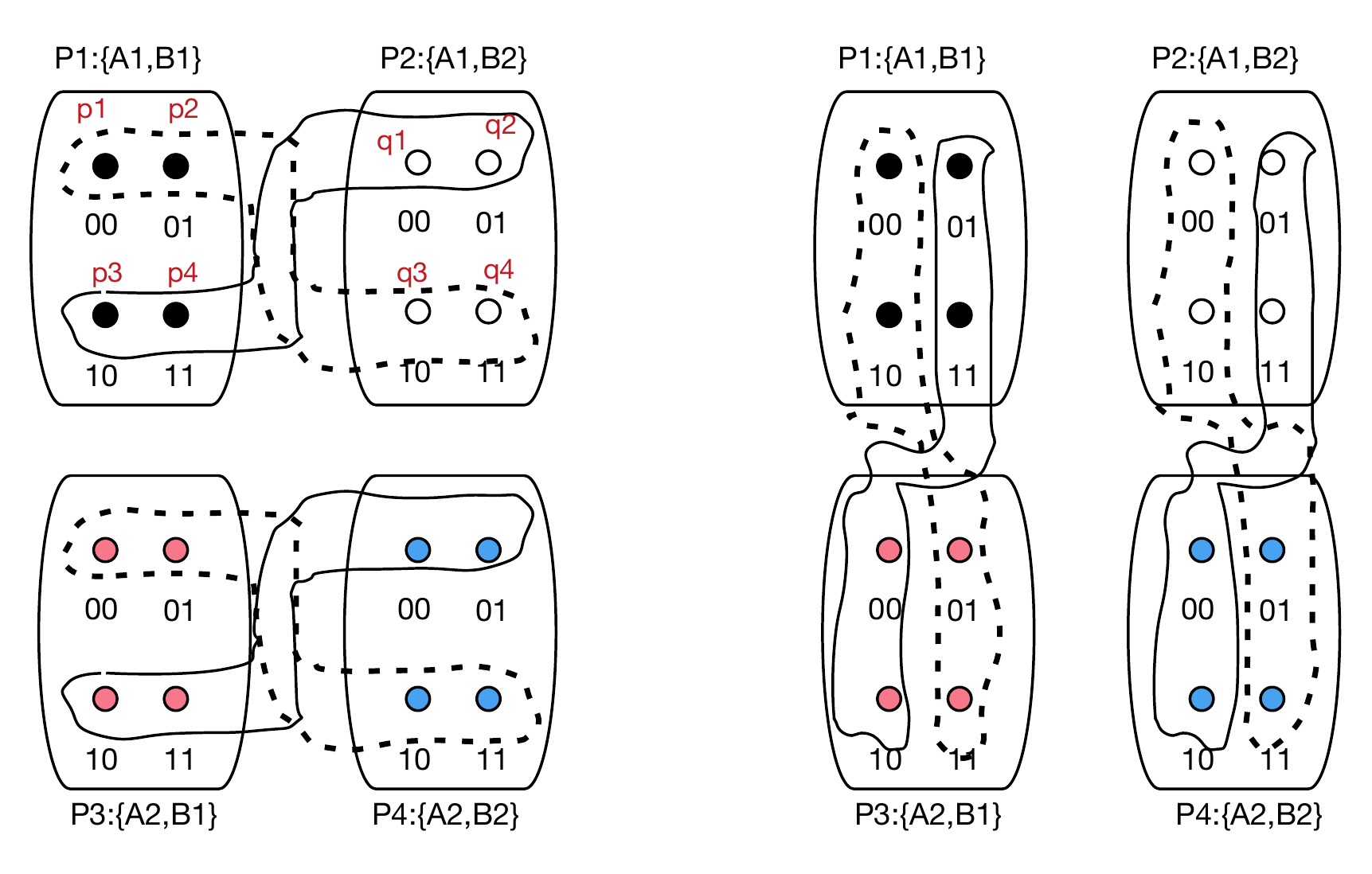}
\caption{Contextuality scenario of the P-program of Figure \ref{pprog:EPR}. This P-program has the cyclic schema depicted in Figure \ref{fig:cyclic}. In total the hypergraph  comprises 12 edges of four events. The nodes in rectangles represent events in a probability distribution returned by a given scope:P1, P2, P3, and P4. 
Note this figure depicts a single hypergraph. Two copies have been made to depict the spanning edges more clearly. This figure corresponds to Figure 7f in \citep{Acin:2015}.}\label{fig:hyperEPR}
\label{fig:X}
\end{figure}

The Bell scenario program syntactically specifies that there should be ``no signalling" between the respective components $A$ and $B$ via the command  \texttt{model({design: `no-signal',P1,P2,P3,P4})}.
This condition imposes constraints on the allowable probabilistic models on the combined hypergraph structure.
Following Definition 3.1.2 in \citep{Acin:2015}, a probabilistic model $p \in \pX{X_A \times X_B}$ is a ``no signalling" model if:
\begin{eqnarray*}
\sum_{w \in e}p(v,w) &=& \sum_{w \in e'}p(v,w), \forall v \in V(X_A), e,e' \in E(X_B) \\
\sum_{w \in e}p(v,w) &=& \sum_{w \in e'}p(v,w), \forall w \in V(X_B), e,e' \in E(X_A) 
\end{eqnarray*}
The probabilistic constraints entailed by this definition will be illustrated in an example to follow.
\citet{Acin:2015}(p45) show that not all probabilistic models of contextuality scenarios composed by a direct product are ``no signalling" models. 
In order to guarantee that all probabilistic models of a combined contextuality scenario are ``no signalling" models, the constituent contextuality scenarios $X_A$ and $X_B$ should be combined by the Foulis-Randall (FR) product denoted $X_{AB} = X_A \FR X_B$.
As with the direct product $X_A \times X_B$ of contextuality scenarios, the vertices of the FR product are defined by  $V(X_A \FR X_B) = V(X_A) \times V(X_B)$. 
It is with respect to the hyperedges that there is a difference between the FR product and the direct product:
\begin{eqnarray*}
X_A \FR X_B &=& E_{A\rightarrow B} \cup E_{B\leftarrow A}
\end{eqnarray*}
where
\begin{eqnarray*}
E_{A\rightarrow B} &\coloneqq & \bigcup_{v \in e_a} \{v\} \times f(v): e_a \in E(X_A), f: e_a \rightarrow E(X_B) \\
E_{A\leftarrow B} &\coloneqq & \bigcup_{w \in e_b} f(w) \times \{w\} \times : e_b \in E(X_B), f: e_b \rightarrow E(X_A) 
\end{eqnarray*}
%FIXME this difference is 
%\comment{PDB}{More here about the FR product}
We are now in  a position to illustrate the semantics of the P-program of Figure \ref{pprog:EPR} by the corresponding contextuality scenario depicted in Figure \ref{fig:hyperEPR}. 
Observe how the FR product produces the extra edges that span the events across measurement contexts labeled P1, P2, P3 and P4 when compared with the direct product hypergraph depicted in Figure \ref{fig:product},
%(Recall that these contexts correspond to the four  syntactic scopes specified in the Bell scenario P-program.)
At first these spanning edges may seem arbitrary, but they happen to guarantee that the allowable probabilistic models  over the composite contextuality scenario  $X_A \FR X_B$ satisfy the ``no signalling" condition \citep{sainz:wolfe:2017}.
By way of illustration, the normalization condition on edges imposes the following constraints (see Figure \ref{fig:hyperEPR}):
\begin{align}
p_1 + p_2 + p_3 + p_4 = 1 \label{eqa}\\
q_1 + q_2 + q_3 + q_4 = 1 \label{eqb}\\
p_1 + p_2 + q_3 + q_4 = 1  \label{eqc} \\
p_3 + p_4 + q_1 + q_2 = 1 \label{eqd}
\end{align}
where $p_i, 1 \leq i \leq 4$ and $q_j, 1 \leq j \leq 4$ denote the probabilities of events in the respective hyperedges. 
A consequence of constraints \eqref{eqa} and \eqref{eqc} is that $p_3 + p_4 = q_3 + q_4$.
When considering the associated outcomes this means 
\[\small \underbrace{p(A1=1 \wedge B1=0)}_{p_3} + \underbrace{p(A1=1 \wedge B1=1)}_{p_4} = \\
\underbrace{p(A1=1 \wedge B2=0)}_{q_3} + \underbrace{p(A1=1 \wedge B2=1)}_{q_4}
\] 
(The preceding is an example of one of the constraints imposed by Definition 3.1.2 in \citep{Acin:2015} as specified above).
In other words, the marginal probability $p(A1=1)$  does not differ across the measurement contexts P1 and P2 specified in the P-program of Figure \ref{pprog:EPR}.
In a similar vein, equations \eqref{eqa} and \eqref{eqd} imply that the marginal probability $p(A1=0)$ does not differ across measurement contexts P1 and P2. 
The stability of marginal probability ensures that ``no signalling" is occurring from component $B$ to component $A$ (see Figure \ref{fig:system}).
%Similar to the order effects contextuality scenario described in the previous section, the spanning hyperedges shown in figure \ref{fig:hyperEPR} imply that the events  with $p_1$ and $p_2$ in the hyperedge corresponding to measurement context P1 are deemed equivalent with the events $q_1$ and $q_2$ respectively in measurement context P2. Similarly the events associated with $p_3$ and $p_4$ are considered equivalent to the events associated with $q_3$ and $q_4$.
In terms of our BOXER BAT example, ``no signalling" implies that the probability of interpretation of the concept BOXER does change whether the priming word for BAT is \emph{ball} ($B1$ - sport sense) or \emph{vampire} ($B2$ - animal sense).
%The equivalent on a physics experiment would be that components A and B are sufficiently spatially separated so that it could be guaranteed that the marginal probability $p(A1)$ is not affected by measurement settings $B1$ and $B2$ used to asked question of component B. (See figure \ref{fig:system}).
\subsection{Bell contextuality scenario with signalling}
Investigations into contextuality in quantum physics  involve the ``no signalling" condition. 
However, in cognitive science and related areas, the situation seems isn't as clear cut.
\citet{dzhafarov:kujala:2015a} argue, for example, that the ``no signalling" condition seems always to be violated in psychological experiments. 
By way of illustration, consider once again the conceptual combination BOXER BAT.
Recall that the ``no signalling" condition entails that the probability of interpretation of the concept BOXER does change whether the priming word for BAT is \emph{ball} ($B1$ - sport sense) or \emph{vampire} ($B2$ - animal sense).
Nor does the probability of interpretation of the concept BAT does change whether the priming word for BOXER is \emph{fighter} ($A1$ - sport sense) or \emph{dog} ($A2$ - animal sense).
However, it is easy to imagine that signalling may be involved in forming an interpretation of BOXER BAT.
For example, \citet{wisniewski:1997} identifies a property-based interpretation of conceptual combinations whereby properties of the modifying concept BOXER apply in some way to the head concept BAT. 
One way to view this kind of interpretation is that a sense of BOXER is first established and then influences the interpretation of the concept BAT. 
In other words, the interpretation of the conceptual combination is formed by processing the combination from left to right.
In relation to the general system depicted in Figure \ref{fig:system}, the preceding situation involves an arrow proceeding from component $A$ to $B$, which represents component $A$ signalling information to component $B$.

We can model Wisniewski's property interpretation by extending the Bell scenario to involve signalling as specified in the P-program shown in Figure \ref{pprog:EPR-signal}.
\begin{figure}
\small
\begin{lstlisting}[ 
           language=Python,
           showspaces=false,
           basicstyle=\ttfamily,
           numbers=left,
           numberstyle=\tiny,
           commentstyle=\color{gray}
        ]
# define the components of the experiment
def A = component(A1,A2)
def B = component(B1,B2)        

var P1= context(){
# signalling: variable B1's outcome is dependent on A1's outcome
     var A1 = flip(0.6)
     var B1 = A1 ? flip(0.8): flip(0.2)     
# declare joint distribution across the variables A1, B1
     var p=[A1,B1]
# flip the dual coins 1000 times to form the joint distribution
     return {Infer({samples:1000},p)}
};
var P2= context(){
     var A1 = flip(0.4)
     var B2 = A1 ? flip(0.3): flip(0.6)  
     var p=[A1,B2]
     return {Infer({samples:1000},p)}
};
var P3= context(){
     var A2 = flip(0.2)
     var B2 = A1 ? flip(0.9): flip(0.2)  
     var p=[A2,B1]
     return {Infer({samples:1000},p)}
};
var P4= context(){
     var A2 = flip(0.4)
     var B2 = A2 ? flip(0.8): flip(0.1)  
     var p=[A2,B2]
     return {Infer({samples:1000},p)}
};
# return a single model
return {model({design: `signal(A->B)',P1,P2,P3,P4})}    
\end{lstlisting}
\vspace{-0.5cm}
\caption{Example P-program specifying a signalling ``Bell" scenario} 
\label{pprog:EPR-signal}
\end{figure}
The signalling from concept $A$ to concept $B$ in a given measurement context is modelled as the outcome of the $B$ coin being dependent on the outcome of the $A$ coin. 
%(Recall that when a coin toss is 1,  the concept has been interpreted in the sense corresponding to the associated random variable). 
Note that signalling does not occur the other way, namely, the probability of interpretation of $A$ does not change according to outcomes measured in relation to component $B$.
This fact allows a more refined understanding of the hypergraph semantics depicted in Figure \ref{fig:hyperEPR} and how these semantics relate to an experimental design which now involves signalling.

\begin{figure}[h!]
\centering
\includegraphics[width=6cm]{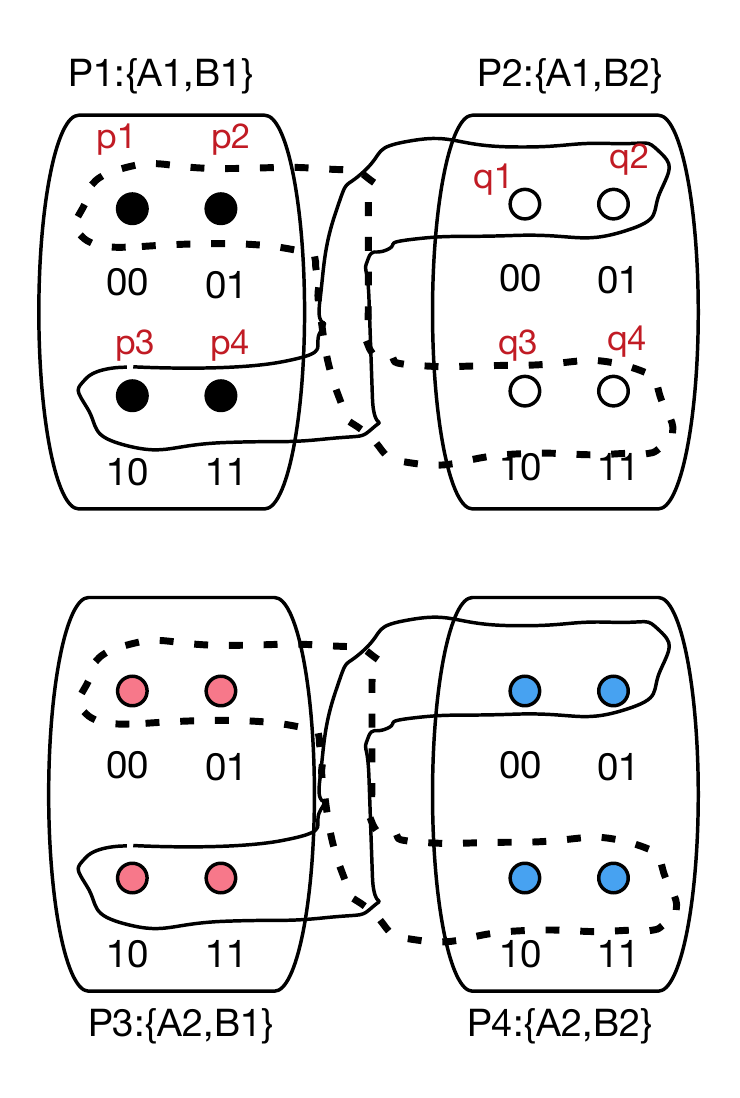}
\caption{Contextuality scenario corresponding to the signalling Bell scenario specified by the P-program in Figure \ref{pprog:EPR-signal}.} \label{fig:hyperEPR-signal}
\end{figure}

In the previous section it was established that the spanning edges in the left hand side of this figure prevents signalling from component $B$ to $A$. 
Conversely, the spanning edges in the right hand side of the figure prevent signalling from $A$ to $B$.
Therefore, the hypergraph semantics of the signalling Bell scenario specified by the program in Figure \ref{pprog:EPR-signal} does not include these right hand side spanning hyperedges.
The resulting hypergraph semantics are depicted as the contextuality scenario shown Figure \ref{fig:hyperEPR-signal} which is the semantics corresponding to the syntax \texttt{model({design: `signal(A->B)',P1,P2,P3,P4})}, where \texttt{A->B} expresses the direction of the signalling between the respective components.
%Note that in contrast to the ``no signalling" Bell scenario P-program (Figure \ref{pprog:EPR}), the syntax \texttt{model({design: `signal',P1,P2,P3,P4})} would produce this particular contextuality scenario which admits signalling from component $A$ to $B$.
Definition \ref{def:contextuality} can then be applied to determine whether a probabilistic model exists in relation to this contextuality scenario.
If not, the signalling system modelled by the P-program in Figure \ref{pprog:EPR-signal} is deemed to be ``strongly contextual".

\section{Discussion}

The aim of this article is to take an algorithmic approach for the development of probabilistic models by providing a high level language that makes it convenient for the modeller to express models of a phenomenon that may be contextual. 
Borrowing from programming language theory, a key feature is the use of syntactic scopes which permits measurement contexts to be specified that correspond to the experimental conditions under which the  phenomenon is being examined. 
The use of syntactic scopes has two consequences. Firstly, random variables local to a scope will be invisible to those local to other scopes.
Secondly, each scope returns a probability distribution as a partial model. 

The first consequence relates to scopes preventing the incorrect overloading of random variables.
This article has attempted to show that the overloading of variables in probabilistic models relates to contextuality namely ``contextual situations are those in which seemingly the same random variable changes its identity depending on the conditions under which it is recorded" \citep{dzhafarov:kujala:2014c}. 

Regarding the second consequence, \citet{Abramsky:2015} discovered that the problem of combining partial models into a single model has an equivalent expression in relational database theory where the problem is to determine whether a universal relation exists for a set of relations such that these relations can be recovered from the universal relation via projection. 
Contextuality occurs when it possible to construct the universal relation.
The question to be addressed, then, is how to determine when it is possible, and when it is not. 

This article proposes hypergraphs as an underlying semantic structure to address this question.
Firstly, an approach developed in relational database theory is used to determine whether the schema of the partial models are acyclic.  If so, the hypergraph is exploited to form a join tree which can compute a single model such that the partial models can be retrieved by appropriately marginalizing this model.

When the schema is cyclic, hypergraphs called ``contextuality scenarios" are formed.
The general picture is the following: Experimental designs are syntactically specified in addition to associated measurement contexts appropriate to the design.
Each component can be translated into a contextuality scenario.
Multipartite composition of these contextuality scenarios yields a single contextuality scenario corresponding to the experimental design.
%Composition, in turn, must be sensitive to a particular design. 
In this article, we illustrated two Bell scenario designs based on whether the ``no signalling" condition holds.
If this condition does hold, then the Foulis-Randall (FR) product can be used to define the composition.
%An important consideration in enforcing the ``no signalling" condition is determining which events are equivalent across measurement contexts. 
%Determining event equivalence can be quite a subtle question related to a specific experimental design.
%One of the advantages of P-programs is that designs can be flagged syntactically and thus allow for equivalent events to be determined by specific syntactic constructs related to a given experimental design.
However, when signalling is permitted, means other than the FR product need to be developed. 
This is an open question which is particularly relevant to psychology experiments where signalling appears to be pervasive.
In this regard, recent work on signalling in Bell scenarios may provide a useful basis for further development \citep{brask:chaves:2017}. 
For example, \citet{brask:chaves:2017} studies relaxations of the ``no signalling" condition where different forms of communication are allowed.
The P-program depicted in Figure \ref{pprog:EPR-signal} modelled one such condition in which outcomes can be uni-directionally communicated between the two components of the assumed model.
Investigating contextuality in the presence of signalling is an important issue for cognitive science and related areas. 
Perhaps surprisingly it is an issue that has received scant attention to date \citep{dzhafarov:kujala:2014b}.
When signalling is not present, it would be interesting to investigate how variations of multipartite composition of contextuality being investigated in physics may inspire new experimental designs outside of physics \citep{sainz:wolfe:2017}. 

Once a contextuality scenario has been constructed for the P-program,
``strong contextuality" occurs when it is not possible to construct a probabilistic model on the underlying hypergraph.
If a probabilistic model on the hypergraph is possible, then the random variables are independent of the measurement contexts.

The motivation for demarcating the problem into acyclic vs. cyclic cases is related to efficiency: The number of variables at the schema level is likely to be much smaller than the number of underlying events, especially when one considers larger scale experiments involving numerous random variables.
This is not withstanding the fact that determining whether there is a global model turns out to be tractable.
Stated more formally, given a contextuality scenario $X$, a linear progam can determine whether strong contextuality holds. (See Proposition 8.1.1 in \citep{Acin:2015}.) 
This theoretical result echoes linear programming solutions which have been found for contextual semantics based on sheaf theory \citep{abramsky:barbosa:mansfiled:2016} and selective influence \citep{dzhafarov:kujala:2012}.

One of the advantages of the hypergraph semantics of contextuality scenarios is that they are general enough to allow contextuality to be investigated in a variety of experimental settings. In the next section we show how contextuality could be investigated in an information fusion setting.

\subsection{The use of P-programs for investigating contextuality in information fusion}

Information fusion refers to the problem of making a judgement about an entity, situation, or event by combining data from multiple sources which are simultaneously presented to a human subject. For example, one source might be an image and another might be a social media post. Fusion allows a much better judgment to be made because it is based on multiple sources of evidence. However, the sources may involve uncertainty, for example, the human subject may not trust the source of a social media post, or the image may appear manipulated. 
As a consequence, a decision of trust may be contextual because a random variable $T$ modelling trust may have different functional identities depending on the source stimulus. 

\begin{figure}[h]
\centering
\includegraphics[width=13cm]{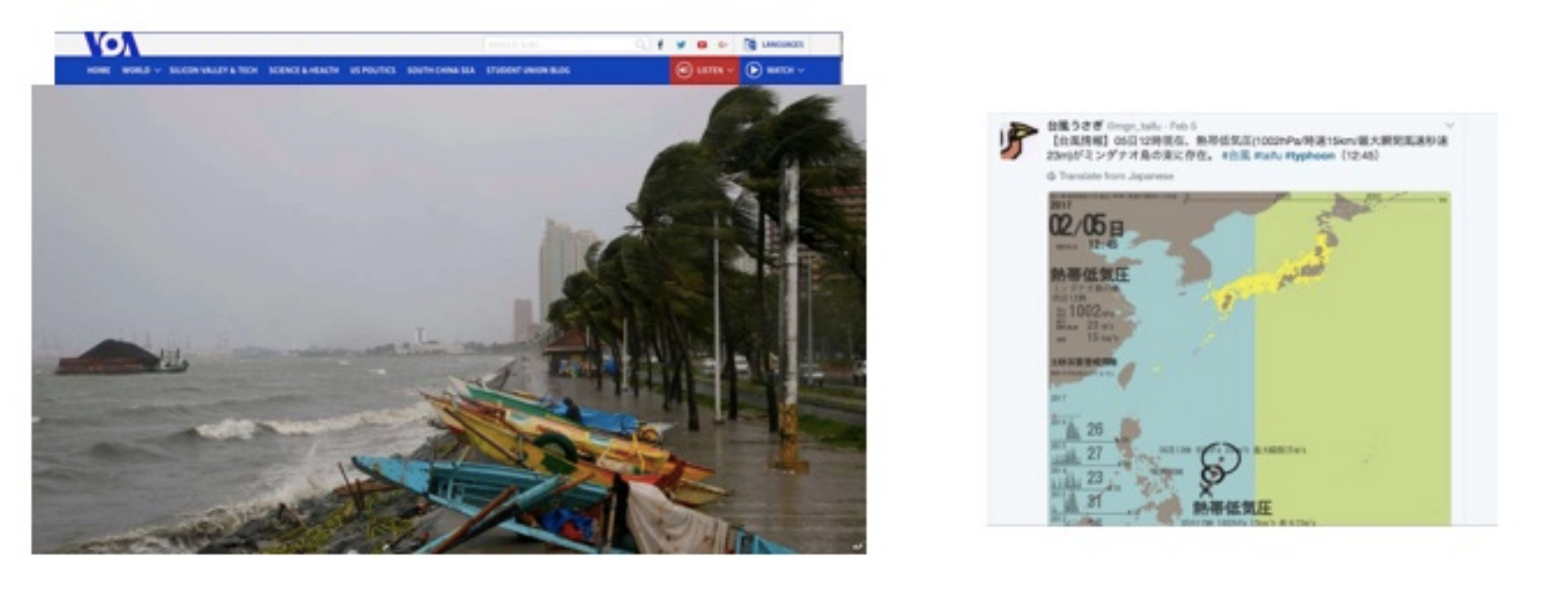}
\caption{Example information fusion scenario. Do the two stimuli pertain to the same event?}
\label{fig:typhoon}
\end{figure}

Let us now sketch how a P-program could be developed to investigate whether trust is contextual.
Firstly, imagine that empirical data is collected from human subjects in an experiment.
For example, subjects could be simultaneously presented with two visual stimuli as is shown in Figure \ref{fig:typhoon}.
The left hand stimulus purports to be a image of a typhoon hitting the Phillipines sourced from an obscure Asian media site.
The right hand stimulus is sourced from Twitter where the language is unfamiliar (Japanese), but the graphic seems to depict a typhoon tracking towards the Phillipines.
The subject must decide if they trust whether the stimuli depict the same event.
Random variables affecting the decision of trust could be defined as follows:
\begin{itemize}
\item  Variables relating to the image: $I_1$ (e.g., ``Do you trust that the image does correspond to the situation described by the text?"),  $I_2$ (e.g., ``Does the image look fake or manipulated in any way?")  
\item Variables related to the tweet: Credibility $S_1$ (e.g., ``Do you trust the source of the tweet to be credible?"), $S_2$ (e.g., ``Do you trust that the tweet corresponds to the situation depicted in the image?").
\end{itemize}
These four variables allow for an experiment in which one variable of each stimulus is measured, thus imply four measurement contexts based on the the following pairs of variables: $\{I_1,S_1\}$,$\{I_1,S_2\}$,$\{I_2,S_1\}$ and 
$\{I_2,S_2\}$.
A between subjects design allows experimental data to be collected in each experimental context meaning a human subject is exposed to only one measurement context in order to counter learning effects.
The corresponding P-program would therefore include four scopes corresponding to these measurement contexts and each scope would return the corresponding partial model based on the data collected in that measurement context. These four partial models correspond to the pairwise distributions: $p(I_1,S_1)$, $\;p(I_1,S_2)$,$\;p(I_2,S_1)$ and $p(I_2,S_2)$.

As this program involves a cyclic schema, the situation is similar to that depicted in Figure \ref{fig:cyclic}.
Therefore, measurement contexts would  be defined around  observations of individual variables $I_1,I_2,S_1,S_2$ using a signalling Bell scenario design. 
As subjects are processing both stimuli simultaneously, it raises the possibility of signalling between the left stimulus (component $A$) and the right stimulus (component $B$).
%Therefore, the corresponding basic contexts would be combined according to a ``within subjects" design. 
%In summary, the P-program would have a very similar for to the one depicted in figure \ref{pprog:EPR}, but where the basic measurement contexts are combined according to a ``within subjects" design.
%It should be noted that even though a within subjects designis specified, signalling mat or may not occur.

%\inlinecomment{FIXME}{more here .. 1) the role of otyer forms of contextuality 2) prospects of using P-programs to model contxtuality in physics in a convenient way. Also psychology, there are many experimental designs, eg Latin square. 3) Possibility to translate experimenst from physics eg KCBS into psychology in order to prove the existence of contextuality in cognition.}

\section{Summary and Future Directions}

%FIXME Bevan to here: rephrase conclusion
The aim of this article to contribute the foundations of a probabilistic programming language that allows exploration of contextuality in wide range of applications relevant to cognitive science and artificial intelligence. 
The core idea is that probabilistic models are specified as a program with associated semantics which are sensitive to contextuality.
The programs feature specific syntactic scopes to specify experimental conditions under which a phenomenon is being examined.
Random variables are declared local to a scope, and hence are not visible to other variables.
In this way, random variables can be safely overloaded which is convenient for developing models whilst the programming semantics, not the modeller, keeps track of the whether functional identities of the random variables are being preserved.

Hypergraphs were proposed as an underlying structure to specify contextually sensitive program semantics.
Firstly, a hypergraph approach developed in relational database theory was used to determine whether the schema of the partial probabilistic models is acyclic.  If so, the hypergraph is exploited to form a join tree which can compute a single model such that the partial models can be retrieved by appropriately marginalizing this model. 
In this case, the phenomenon is non-contextual.
When the schema is cyclic, the phenomenon may or may not be contextual.
For the cyclic case a hypergraph called a ``contextuality scenario" is formed. 
%The general picture is the following: each syntactic scope will translate into an edge of the hypergraph. The vertices in the edges correspond to events. Events are outcomes that can be observed in the measurement context specified by the associated syntactic scope. An important consideration in contextuality scenarios is determining which events are equivalent across measurement contexts. This can be a quite subtle question related to a specific experimental design. One of the advantages of P-programs is that designs can be flagged syntactically and thus allow for equivalent events to be determined by syntactical analysis of the program.
``Strong contextuality" occurs when it is not possible to construct a probabilistic model on the hypergraph.
If it is possible, then each such model is a candidate global model and the phenomenon is non-contextual.
Further research could be directed at refining the semantics to admit different types of contextuality \citep{abramsky:brandenburger:2011,Acin:2015}, as well as experimental designs based on different variations of signalling \citep{brask:chaves:2017,wittek:acin:2017}. 
%This is important as many experiments in cognitive science and related areas are likely to involve signalling.

%Another advantage of using a programming approach to develop probabilistic models is that experimental designs can be syntactically specified in a modular way in order to conveniently express more complicated designs. 
%In this way a wide variety of experimental designs across fields can potentially be catered for. In this article, ``between" and ``within" subjects designs were illustrated. Further research, could be directed towards investigating how P-programs can express a wide range of experimental designs to suit a diversity of experimental conditions. 
Just like higher level programming languages, such as functional programming, provided a convenient means for harnessing the power of the lambda-calculus,  
P-programs aim to advance the understanding of contextuality, by providing a convenient means for harnessing the power of contextual semantics.
%This has the power for making quantum information processing more powerful, more quickly.
As P-programs are algorithmic, future work could provide syntax to specify the temporal flow of actions using control structures akin to those used in high level programming languages. 
This feature allows measurements with some causal structure, which is an important topic in cognitive psychology where Bayesian models are often used.

Finally, the overarching aim of this article is to raise awareness of contextuality beyond quantum physics and to contribute formal methods to detect its presence in the form of a convenient programming language.

%The ultimate goal is to deliver an expressive convenient language for modelling contextual phenomena in a wide variety of applications relevant to cognitive science and artificial intelligence. 
\subsubsection*{Acknowledgements}
Thanks to the three anonymous reviewers, Bevan Koopman and Ana Belen Sainz for their constructive input and suggestions. This research was supported by the Asian Office of Aerospace Research and Development (AOARD) grant: FA2386-17-1-4016 
%\bibliographystyle{apa}
%\bibliography{new1}

\begin{thebibliography}{}

\bibitem[\protect\astroncite{Abramsky}{2015}]{Abramsky:2015}
Abramsky, S. (2015).
\newblock Contextual semantics: From quantum mechanics to logic, databases,
  constraints, and complexity.
\newblock {\em Bulletin of EATCS}, 2(113).

\bibitem[\protect\astroncite{Abramsky
  et~al.}{2016}]{abramsky:barbosa:mansfiled:2016}
Abramsky, S., Barbosa, R., and Mansfield, S. (2016).
\newblock Quantifying contextuality via linear programming.
\newblock In {\em Proceedings of the 13th International Conference on Quantum
  Physics and Logic (QPL 2016)}.

\bibitem[\protect\astroncite{Abramsky and
  Brandenburger}{2011}]{abramsky:brandenburger:2011}
Abramsky, S. and Brandenburger, A. (2011).
\newblock The sheaf-theoretic structure of non-locality and contextuality.
\newblock {\em New Journal of Physics}, 13(113036).

\bibitem[\protect\astroncite{Acacio De~Barros and Oas}{2015}]{barros:oas:2015}
Acacio De~Barros, J. and Oas, G. (2015).
\newblock Some examples of contextuality in physics: {I}mplications to quantum
  cognition.
\newblock arXiv:1512.00033.

\bibitem[\protect\astroncite{Acin et~al.}{2015}]{Acin:2015}
Acin, A., Fritz, T., Leverrier, A., and Sainz, A. (2015).
\newblock A combinatorial apporoach to nonlocality and contextuality.
\newblock {\em Communications in Mathematical Physics}, 334:533--628.

\bibitem[\protect\astroncite{Aerts et~al.}{2014}]{Aerts:2014}
Aerts, D., Gabora, L., and Sozzo, S. (2014).
\newblock Concept combination, entangled measurements, and prototype theory.
\newblock {\em Topics in Cognitive Science}, 6:129--137.

\bibitem[\protect\astroncite{Aerts and Sozzo}{2014}]{aerts:sozzo:2014}
Aerts, D. and Sozzo, S. (2014).
\newblock Quantum entanglement in concept combinations.
\newblock {\em International Journal of Theoretical Physics}, 53:3587--3603.

\bibitem[\protect\astroncite{Asano et~al.}{2014}]{khrennikov:2014}
Asano, M., Hashimoto, T., Khrennikov, A., Ohya, M., and Tanaka, Y. (2014).
\newblock Violation of contextual generalization of the {L}eggett--{G}arg
  inequality for recognition of ambiguous figures.
\newblock {\em Physica Scripta}, 2014(T163):014006.

\bibitem[\protect\astroncite{Atmanspacher and
  Filk}{2010}]{atmanspacher:filk:2010}
Atmanspacher, H. and Filk, T. (2010).
\newblock A proposed test of temporal nonlocality in bistable perception.
\newblock {\em Journal of Mathematical Psychology}, 54:314--321.

\bibitem[\protect\astroncite{Brask and Chaves}{2017}]{brask:chaves:2017}
Brask, J.~B. and Chaves, R. (2017).
\newblock Bell scenarios with communication.
\newblock {\em Journal of Physics A: Mathematical and Theoretical},
  50(9):094001.

\bibitem[\protect\astroncite{Bruza and Abramsky}{2016}]{bruza:abramsky:2017}
Bruza, P. and Abramsky, S. (2016).
\newblock Probabilistic programs: Contextuality and relational database theory.
\newblock In Acacio De~Barros, J., Coecke, B., and Pothos, E., editors, {\em
  Quantum Interaction: 10th International Conference (QI'2016)}, {Lecture Notes
  in Computer Science}. Springer (In Press).

\bibitem[\protect\astroncite{Bruza et~al.}{2015}]{bruza:kitto:ramm:sitbon:2015}
Bruza, P., Kitto, K., Ramm, B., and Sitbon, L. (2015).
\newblock A probabilistic framework for analysing the compositionality of
  conceptual combinations.
\newblock {\em Journal of Mathematical Psychology}, 67:26--38.

\bibitem[\protect\astroncite{Bruza}{2016}]{bruza:2016}
Bruza, P.~D. (2016).
\newblock Syntax and operational semantics of a probabilistic programming
  language with scopes.
\newblock Journal of Mathematical Psychology 10.1016/j.jmp.2016.06.006 (In
  press).

\bibitem[\protect\astroncite{Clauser and Horne}{1974}]{clauser:horne:74}
Clauser, J. and Horne, M. (1974).
\newblock Experimental consequences of objective local theories.
\newblock {\em Physical Review D}, 10(2):526--535.

\bibitem[\protect\astroncite{Curchod et~al.}{2017}]{wittek:acin:2017}
Curchod, F., Johansson, M., Augusiak, R., Hoban, M., Wittek, P., and Acin, A.
  (2017).
\newblock Unbounded randomness certification using sequences of measurements.
\newblock arXiv:1510.03394v2.

\bibitem[\protect\astroncite{Dzhafarov and
  Kujala}{2014a}]{dzhafarov:kujala:2014c}
Dzhafarov, E. and Kujala, J. (2014a).
\newblock Contextuality is about identity of random variables.
\newblock {\em Physica Scripta}, T163(014009).

\bibitem[\protect\astroncite{Dzhafarov and
  Kujala}{2014b}]{dzhafarov:kujala:2014b}
Dzhafarov, E. and Kujala, J. (2014b).
\newblock Embedding quantum into classical: Contextualization vs
  conditionalization.
\newblock {\em {PLoS ONE}}, 9(3):e92818.

\bibitem[\protect\astroncite{Dzhafarov and
  Kujala}{2015}]{dzhafarov:kujala:2015a}
Dzhafarov, E. and Kujala, J. (2015).
\newblock Probabilistic contextuality in {EPR/Bohm-type} systems with signaling
  allowed.
\newblock In Dzhafarov, E., editor, {\em Contextuality from Quantum Physics to
  Psychology.}, chapter~12, pages 287--308. World Scientific Press.

\bibitem[\protect\astroncite{Dzhafarov
  et~al.}{2015}]{dzhafarov:zhang:kujala:2015}
Dzhafarov, E., Zhang, R., and Kujala, J. (2015).
\newblock Is there contextuality in behavioral and social systems?
\newblock {\em Philosophical Transactions of the Royal Society A},
  374(20150099).

\bibitem[\protect\astroncite{Dzhafarov and
  Kujala}{2012}]{dzhafarov:kujala:2012}
Dzhafarov, R. and Kujala, J. (2012).
\newblock Selectivity in probabilistic causality: Where psychology runs into
  quantum physics.
\newblock {\em Journal of Mathematical Psychology}, 56(1):54--63.

\bibitem[\protect\astroncite{Gabora and Aerts}{2002}]{Article:02:Aerts:QM}
Gabora, L. and Aerts, D. (2002).
\newblock {Contextualizing concepts using a mathematical generalization of the
  quantum formalism}.
\newblock {\em Journal of Experimental and Theoretical Artificial
  Intelligence}, 14:327--358.

\bibitem[\protect\astroncite{Goodman and
  Stuhlm\"{u}ller}{2014}]{goodman:stuhlmuller:2014}
Goodman, N.~D. and Stuhlm\"{u}ller, A. (2014).
\newblock {The Design and Implementation of Probabilistic Programming
  Languages}.
\newblock \url{http://dippl.org}.
\newblock Accessed: 2017-9-14.

\bibitem[\protect\astroncite{Goodman and Tenenbaum}{2016}]{probmods2}
Goodman, N.~D. and Tenenbaum, J.~B. (2016).
\newblock {Probabilistic Models of Cognition}.
\newblock \url{http://probmods.org/v2}.
\newblock Accessed: 2017-6-5.

\bibitem[\protect\astroncite{Gordon et~al.}{2014}]{gordon:henzinger:2014}
Gordon, A., Henzinger, T., Nori, A., and Rajamani, S. (2014).
\newblock Probabilistic programming.
\newblock In {\em Proceedings of the on Future of Software Engineering (FOSE
  2014)}, pages 167--181. ACM Press.

\bibitem[\protect\astroncite{Gronchi and
  Strambini}{2016}]{gronchi:strambini:2017}
Gronchi, G. and Strambini, E. (2016).
\newblock Quantum cognition and {B}ell's inequality: A model for probabilistic
  judgment bias.
\newblock {\em Journal of Mathematical Psychology}, 78:65--75.

\bibitem[\protect\astroncite{Gyssens and
  Paredaens}{1984}]{gyssens:paradaens:84}
Gyssens, M. and Paredaens, J. (1984).
\newblock A decomposition methodology for cyclic databases.
\newblock In {\em Advances in Database Theory}, volume~2, pages 85--122.
  Springer.

\bibitem[\protect\astroncite{Henson and Sainz}{2015}]{henson:2015}
Henson, J. and Sainz, A. (2015).
\newblock Macroscopic noncontextuality as a principle for almost-quantum
  correlations.
\newblock {\em Phyical Review A}, 91:042114.

\bibitem[\protect\astroncite{Howard et~al.}{2014}]{howard:2014:contextuality}
Howard, M., Wallman, J., Veitch, V., and Emerson, J. (2014).
\newblock Contextuality supplies the magic for quantum computation.
\newblock {\em Nature}, 510(7505):351--355.

\bibitem[\protect\astroncite{Kochen and Specker}{1967}]{kochen:specker:67}
Kochen, S. and Specker, E. (1967).
\newblock The problem of hidden variables in quantum mechanics.
\newblock {\em Journal of Mathematics and Mechanics}, 17(59).

\bibitem[\protect\astroncite{Liu et~al.}{2011}]{liu:yue:li:2011}
Liu, W., Yue, K., and Li, W. (2011).
\newblock Constructing the {B}ayesian network structure from dependencies
  implied in multiple relational schemas.
\newblock {\em Expert systems with Applications}, 38:7123--7134.

\bibitem[\protect\astroncite{Pelletier}{1994}]{pelletier:1994}
Pelletier, J. (1994).
\newblock The principle of semantic compositionality.
\newblock {\em Topoi}, 13:11--24.

\bibitem[\protect\astroncite{Sainz and Wolfe}{2017}]{sainz:wolfe:2017}
Sainz, A. and Wolfe, E. (2017).
\newblock Multipartite composition of contextuality scenarios.
\newblock arXiv:1701.05171 [quant-ph].

\bibitem[\protect\astroncite{Wisniewski}{1997}]{wisniewski:1997}
Wisniewski, E.~J. (1997).
\newblock When concepts combine.
\newblock {\em Psychonomic Bulletin and Review}, 42(2):167--183.

\bibitem[\protect\astroncite{Wong}{1997}]{wong:1997}
Wong, S. (1997).
\newblock An extended relational model for probabilistic reasoning.
\newblock {\em Journal of Intelligent Information Systems}, 9:181--202.

\bibitem[\protect\astroncite{Wong}{2001}]{wong:2001}
Wong, S. (2001).
\newblock The relational structure of belief networks.
\newblock {\em Journal of Intelligent Information Systems}, 16:117--148.

\bibitem[\protect\astroncite{Wulf and Shaw}{1973}]{wulf:shaw:1973}
Wulf, W. and Shaw, M. (1973).
\newblock Global variable considered harmful.
\newblock {\em SIGPLAN Notices}, 8(2):80--86.

\bibitem[\protect\astroncite{Zhang and Dzhafarov}{2017}]{zhang:dzhafarov:2017}
Zhang, R. and Dzhafarov, E.~S. (2017).
\newblock Testing contextuality in cyclic psychophysical systems of high ranks.
\newblock In Acacio De~Barros, J., Coecke, B., and Pothos, E., editors, {\em
  Quantum Interaction: 10th International Conference (QI'2016)}, {Lecture Notes
  in Computer Science}. Springer (In Press).

\end{thebibliography}

\end{document}